\documentclass[letterpaper, 10 pt, conference]{ieeeconf}

\IEEEoverridecommandlockouts                              
\usepackage{epsfig}
\usepackage{epstopdf}
\usepackage{graphics}
\usepackage{upgreek}
\usepackage{amssymb}
\usepackage{amsfonts}
\usepackage{amsmath}
\usepackage{refstyle}
\usepackage{xcolor}
\usepackage{subfigure}
\usepackage{todonotes}
\usepackage{mathtools}
\usepackage{cite}
\usepackage{balance}
\usepackage{mathrsfs}
\usepackage{lipsum}
\usepackage{multicol}
\usepackage{soul}
\usepackage{siunitx}
\usepackage{amsmath}
\usepackage{algorithm,algpseudocode}
\usepackage{amsfonts}
\usepackage{booktabs}
\usepackage{optidef}
\usepackage{siunitx}

\algnewcommand\algorithmicforeach{\textbf{for each}}
\algdef{S}[FOR]{ForEach}[1]{\algorithmicforeach\ #1\ \algorithmicdo}

\title{\LARGE \bf Robotic Navigation Autonomy for Subretinal Injection \\ via Intelligent Real-Time Virtual iOCT Volume Slicing}

\author{Shervin Dehghani$^{1, 4}$, Michael Sommersperger$^{1}$, Peiyao Zhang$^{4}$, Alejandro Martin-Gomez$^{4}$,\\ Benjamin Busam$^{1},$ 
Peter Gehlbach$^{3}$, Nassir Navab$^{5}$, M. Ali Nasseri$^{1,2}$ and Iulian Iordachita$^{4}$%
\thanks{This work was supported by U.S. National Institutes of Health under
the grants number 2R01EB023943-04A1 and 1R01 EB025883-01A1, and
partially by JHU internal funds.}
\thanks{Corresponding author: Shervin Dehghani (shervin.dehghani@tum.de)}
\thanks{$^{1}$ S. Dehghani, M. Sommersperger, B. Busam and M. Ali Nasseri are with Department of Computer Science, Technische Universit\"{a}t M\"{u}nchen, M\"{u}nchen 85748 Germany. 
}%
\thanks{$^{2}$ M. Ali Nasseri is with Augenklinik und Poliklinik, Klinikum rechts der Isar der Technische Universit\"{a}t M\"{u}nchen, M\"{u}nchen 81675 Germany.
        }%
\thanks{$^{3}$P. Gehlbach is with Wilmer Eye Institute, Johns Hopkins Hospital, Baltimore, MD, USA.
        }%
\thanks{$^{4}$S. Dehghani, Peiyao Zhang, Alejandro Martin-Gomez and I. Iordachita are with Laboratory for Computational Sensing and Robotics, Johns Hopkins University, Baltimore, MD, USA.
}%
\thanks{$^{5}$N. Navab is a full professor and head of the Chair for Computer Aided Medical Procedures \&
Augmented Reality, Technical University of Munich, 85748 Munich, Germany,
and an adjunct professor at the Whiting School of Engineering, Johns Hopkins University,
Baltimore, MD, USA.}%
}
\DeclarePairedDelimiterX{\norm}[1]{\lVert}{\rVert}{#1}
\begin{document}
\definecolor{red}{rgb}{1,0,0}
\newcommand{\later}[3]{\textcolor{red}{{#3}}}

\newcommand*{\vertbar}{\rule[-1ex]{0.5pt}{2.5ex}}
\newcommand*{\horzbar}{\rule[.5ex]{2.5ex}{0.5pt}}

\maketitle


\begin{abstract}
In the last decade, various robotic platforms have been introduced that could support delicate retinal surgeries.
Concurrently, to provide semantic understanding of the surgical area, recent advances have enabled microscope-integrated intraoperative Optical Coherent Tomography (iOCT) with high-resolution 3D imaging at near video rate.
The combination of robotics and semantic understanding enables task autonomy in robotic retinal surgery, such as for subretinal injection.
This procedure requires precise needle insertion for best treatment outcomes.
However, merging robotic systems with iOCT introduces new challenges.
These include, but are not limited to high demands on data processing rates and dynamic registration of these systems during the procedure.
In this work, we propose a framework for autonomous robotic navigation for subretinal injection, based on intelligent real-time processing of iOCT volumes.
Our method consists of an instrument pose estimation method, an online registration between the robotic and the iOCT system, and trajectory planning tailored for navigation to an injection target.
We also introduce \textit{intelligent virtual B-scans}, a volume slicing approach for rapid instrument pose estimation, which is enabled by Convolutional Neural Networks (CNNs).
Our experiments on ex-vivo porcine eyes demonstrate the precision and repeatability of the method.
Finally, we discuss identified challenges in this work and suggest potential solutions to further the development of such systems.
\end{abstract}

\bstctlcite{IEEEexample:BSTcontrol}



\section{Introduction}
Over the past decade, various robotic systems have been introduced to enhance ophthalmic surgery~\cite{ullrich2013mobility,rahimy2013robot,gijbels2014experimental,nasseri2013introduction,molaei2017toward, he2012toward}. 
The utility of such systems is particularly evident in the treatment of complex retinal diseases that require delicate navigation and precise positioning of instruments, exceeding human capabilities in manual surgery.
\begin{figure}[h]
    \centering
    \includegraphics[width=0.8\linewidth]{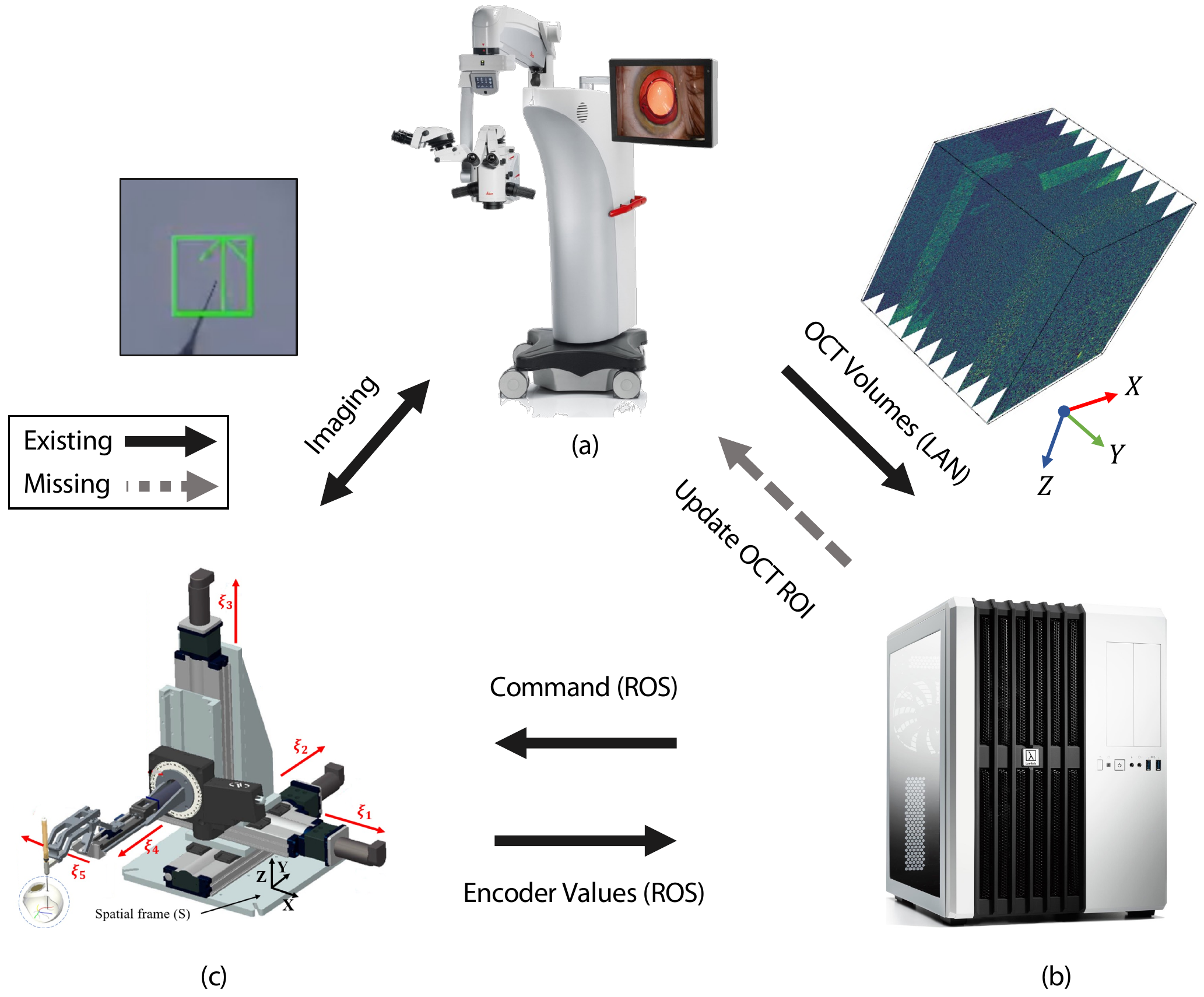}
        \caption{Overview of the proposed system, consisting of an iOCT Microscope (a), which is used for volume acquisition and data transmission to the computation unit (b). A following step enables the surgeon can choose a target point for applying an injection. All the necessary computations are applied and transmitted to the robot (c) to execute the injection.}
    \label{fig:overview}
\end{figure}
The treatment of age-related macular degeneration (AMD) is an example of a potential motivation to employ precise robotic systems.
AMD is a disease that in its advanced neovascular stage, is characterized by leakage of fluid and blood near the macula. The leakage lead to irreversible damage to the retinal cells, and loss of vision.
AMD is considered the leading cause of visual impairment in patients over age 65 in developed countries \cite{Wong2020} and is predicted to affect 288 million people by 2040 \cite{Casten2016}.
The current standard of care for AMD is intravitreous injection of anti-VEGF drugs \cite{Finger2020}. 
This requires repeated treatments to delay progression, but does not lead to a cure.


Novel advancements include, but are not limited to stem cell therapy \cite{Wang2020}, retinal pigment epithelium (RPE) cell transplants \cite{Zarbin2019}, and gene editing technology \cite{Chung2020}.
These may eventually offer an efficient treatment of AMD with a single intervention, enabled by precise subretinal delivery of the therapeutic agent to the potential space between the photoreceptors and the RPE-Bruch's membrane complex \cite{Karampelas1256}.
For this reason, robotic approaches are envisioned to precisely guide a microsurgical injection needle into the subretinal space, in order to safely deliver a therapeutic payload.

To aid surgical visualization and to reveal retinal layer structures, intraoperative Optical Coherence Tomography (iOCT) has been integrated into the surgical setup in the last decade. 
This modality enables 2D and 3D micrometer-resolution imaging of the surgical area and can provide video-rate volumetric imaging due to novel advances in scanning technology \cite{Carrasco-Zevallos2014,Carrasco-Zevallos2018}.
Studies have shown that iOCT has the potential to enable targeted delivery of therapeutic agents into the subretinal space \cite{Xue2017}. 
Prior work has combined iOCT imaging with a manually controlled robotic system \cite{Ladha2021,Nasseri2017}, confirming the advantages of employing a robotic system for subretinal injection, such as reduced tremor, higher bleb formation and reduction of reflux.
\begin{figure}[h!]
    \centering
    \includegraphics[width=0.8\linewidth]{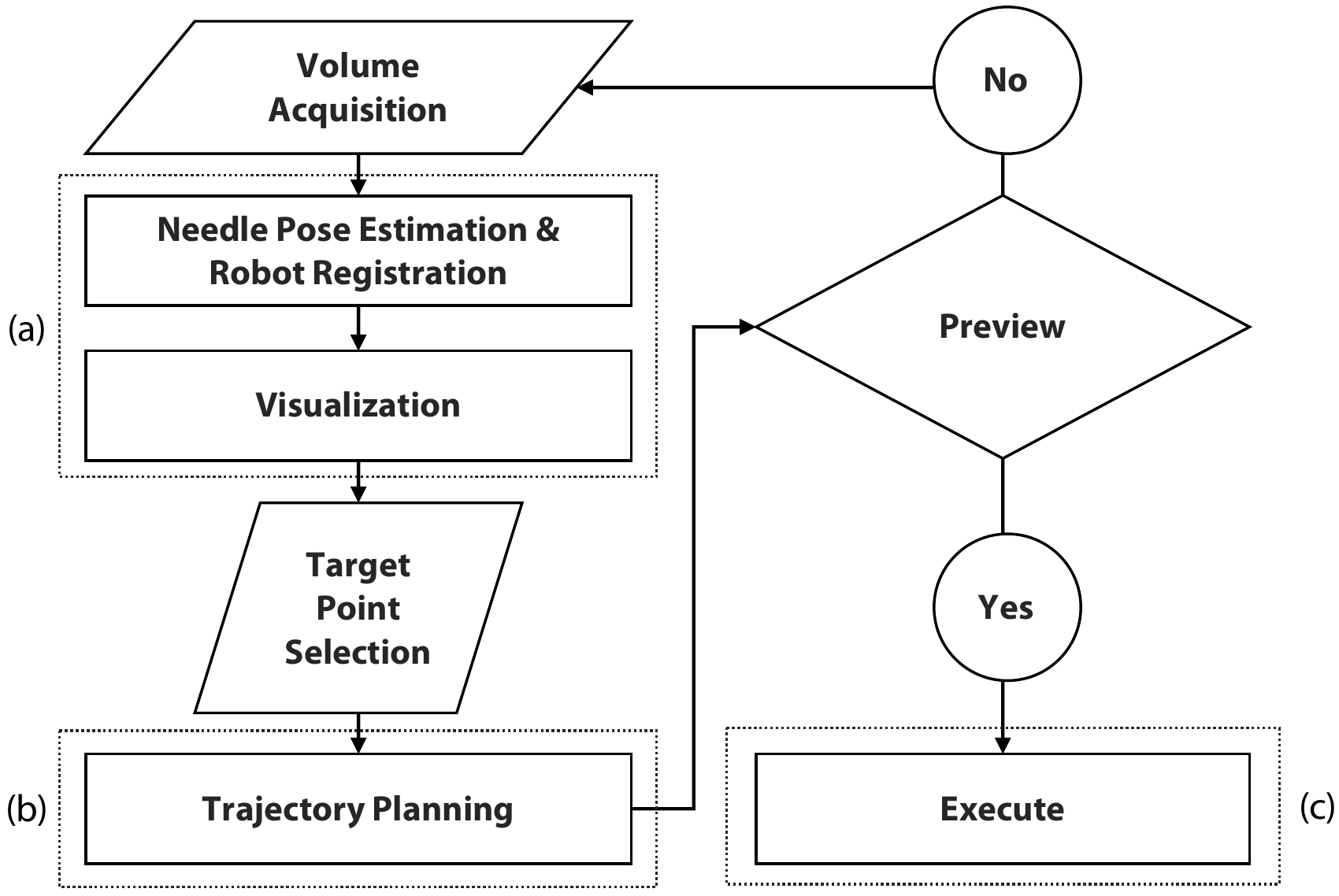}
        \caption{The proposed workflow follows volume processing and visualization (a), planning (b) and execution (c) stages.}
    \label{fig:workflow}
\end{figure}
In this work, we present a system for autonomous robot navigation, which is tailored for the use case of subretinal injection.
The contributions of this work are as follows:
\textbf{1)} We present a framework that merges a robotic system with an iOCT-integrated microscope (Fig.~\ref{fig:overview}), to create a surgeon-in-the-loop workflow (Fig.~\ref{fig:workflow}).
The robot motion is enabled by real-time processing of volumetric iOCT data, in an effort to navigate a surgical needle to a subretinal injection target chosen by the surgeon.
The proposed workflow includes data acquisition, visualization, and selection of a target point within the acquired volume, followed by automatic volume processing and robot navigation.
The final trajectory planning and motion is based on an online registration between the robotic instrument and the iOCT volume.
\textbf{2)} We introduce a novel concept of \textit{intelligent virtual B-scans} as automatically selected instrument-aligned slices, reducing the complexity of the 3D iOCT data.
In our framework this facilitates the integration of convolutional neural networks (CNNs) for instrument pose estimation and trajectory planning in real-time.
We evaluate the proposed framework by first performing targeted navigation to arbitrary points in the volume, and then by performing needle insertions in ex-vivo porcine eyes.
The results show high accuracy and repeatability, encouraging further iOCT-guided robotic vitreoretinal applications. Our results also demonstrate the potential of \textit{virtual B-scans} for real-time volumetric iOCT processing.

\section{Related work}
\label{sec:Relatedwork}

Prior work has addressed components of our framework, such as instrument pose estimation, OCT B-scan segmentation, and the calibration of an iOCT system to a robotic setup.
\textit{Needle pose estimation} from iOCT volumes has been proposed in various works \cite{Zhou2017Beveled,Zhou2018,Weiss2018}. 
Such initial works addressing 5-degree-of-freedom (DoF) tool tracking methods were latter extended to a 6-DoF localization approach \cite{Zhou2019}.
While these methods have shown a high accuracy on instrument detection and pose estimation, they are limited to tracking the needle above the retina and are based on processing of the entire iOCT volumes. This in turn poses constraints on the integrated processing algorithms to cope with the high data throughput of high-speed iOCT systems.
More advanced detection and segmentation algorithms are still required to improve the robustness and generalizabality of such instrument tracking.

With the technical advances in recent years, deep learning approaches for the \textit{segmentation of OCT B-scans} have shown high success rates \cite{VIEDMA2022}. 
Convolutional neural network (CNN) architectures have shown promising results for the segmentation of retinal layers \cite{Roy2017}.
Recently, 3D network architectures \cite{Mukherjee2022} have been developed and specifically tailored to segment whole OCT volumes in the presence of clinical pathology related to AMD.
Most of these works, however, focus on the segmentation of retinal structures in diagnostic OCT, with higher image quality as compared to current iOCT imaging, and do not include surgical instruments and their related artifacts, such as shadows obscuring the retinal anatomy or mirroring artifacts.
\begin{figure}[h!]
    \centering
    \includegraphics[width=0.8\linewidth]{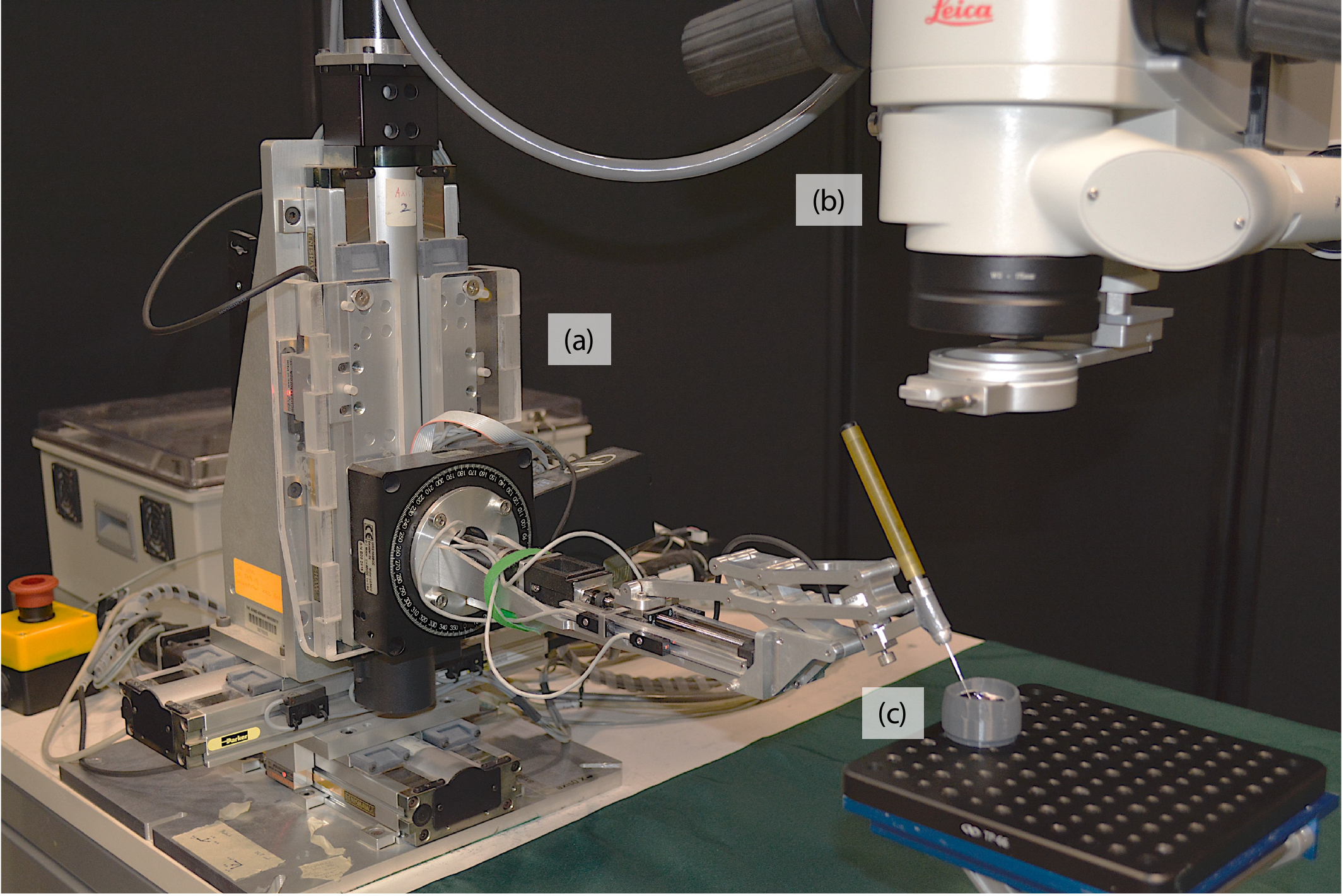}
        \caption{Setup overview. (a) Steady-Hand Eye-robot. (b) Leica iOCT-integrated Microscope. (c) Ex-vivo pig eye.}
    \label{fig:setup}
\end{figure}
To our knowledge there are only a few works addressing advanced instrument segmentation in iOCT imaging, such as the development of a fully convolutional network for the instrument segmentation in iOCT B-scans \cite{Zhou2017NeedleSeg} and instrument segmentation in axial iOCT projection images used for real-time visualization of 4D iOCT \cite{Weiss2020}.
A joint instrument and retinal layer segmentation from iOCT imaging for distance estimation between the instrument and selected retinal layers was proposed only recently \cite{Sommersperger2021}.
Similar to dedicated iOCT segmentation algorithms, methods on the calibration of a volumetric iOCT and a robotic system are to date, sparse and limited to expensive 3D instrument segmentation \cite{Zhou2018Calib}.

\section{Method}
\subsection{Overview}
\label{Sec:MethodOverview}
In this work we address the composition of a system that comprises the necessary components for robotic navigation based on iOCT, which includes rapid volume processing, instrument pose estimation, registration, and trajectory planning.
The setup connects a robotic system with an iOCT-integrated microscope to meet the real-time requirements of intraoperative applications (Fig.~\ref{fig:setup}), while keeping the human in the loop.
The designed workflow consists of a sequence of steps: first, an iOCT volume is acquired. The required processing and instrument pose estimation is applied, and the volume is visualized (Fig.~\ref{fig:workflow} (a)). 
The surgeon then chooses a 3D target point for subretinal injection in the volume. 
Subsequently, a trajectory for navigating to the specified target location is estimated (Fig.~\ref{fig:workflow} (b)). Upon approval, the command is transmitted to the robot for execution (Fig.~\ref{fig:workflow} (c)).
The components of our workflow are specifically tailored to the use-case of subretinal injection.

To perform a successful robotic injection, two features need to be extracted from an input volume: \textbf{1)} the needle pose w.r.t the iOCT volume and \textbf{2)} the optimal trajectory to guide the instrument to a selected target point for subretinal injection.
While 3D learning-based methods show promising results for high-resolution diagnostic OCT, they are not yet suitable for real-time 3D analysis.
Additionally, because of the non-uniform voxel-spacing of the acquired iOCT volumes, conventional 3D CNN-based methods are not directly applicable.
In our approach, we therefore decompose the problem of needle pose estimation into two consecutive 2D segmentation tasks.
We introduce \textit{intelligent virtual B-scans} to segment only selected regions extracted in a rapid volume-processing step.
This is further discussed in Sec.~\ref{Sec:MethodAxial} and Sec.~\ref{sec:MethodBScan}. 
Afterwards, in Sec.~\ref{sec:MethodRegistration} the needle pose estimation and the registration between the robot and volume are described.
Finally, the trajectory for executing the injection at the given 3D target point is calculated and transmitted to the robot (Sec.~\ref{sec:TrajectoryPlanning}).

\subsection{Axial Projection Image}
\label{Sec:MethodAxial}
\begin{figure}[H]
    \centering
    \includegraphics[width=0.5\linewidth]{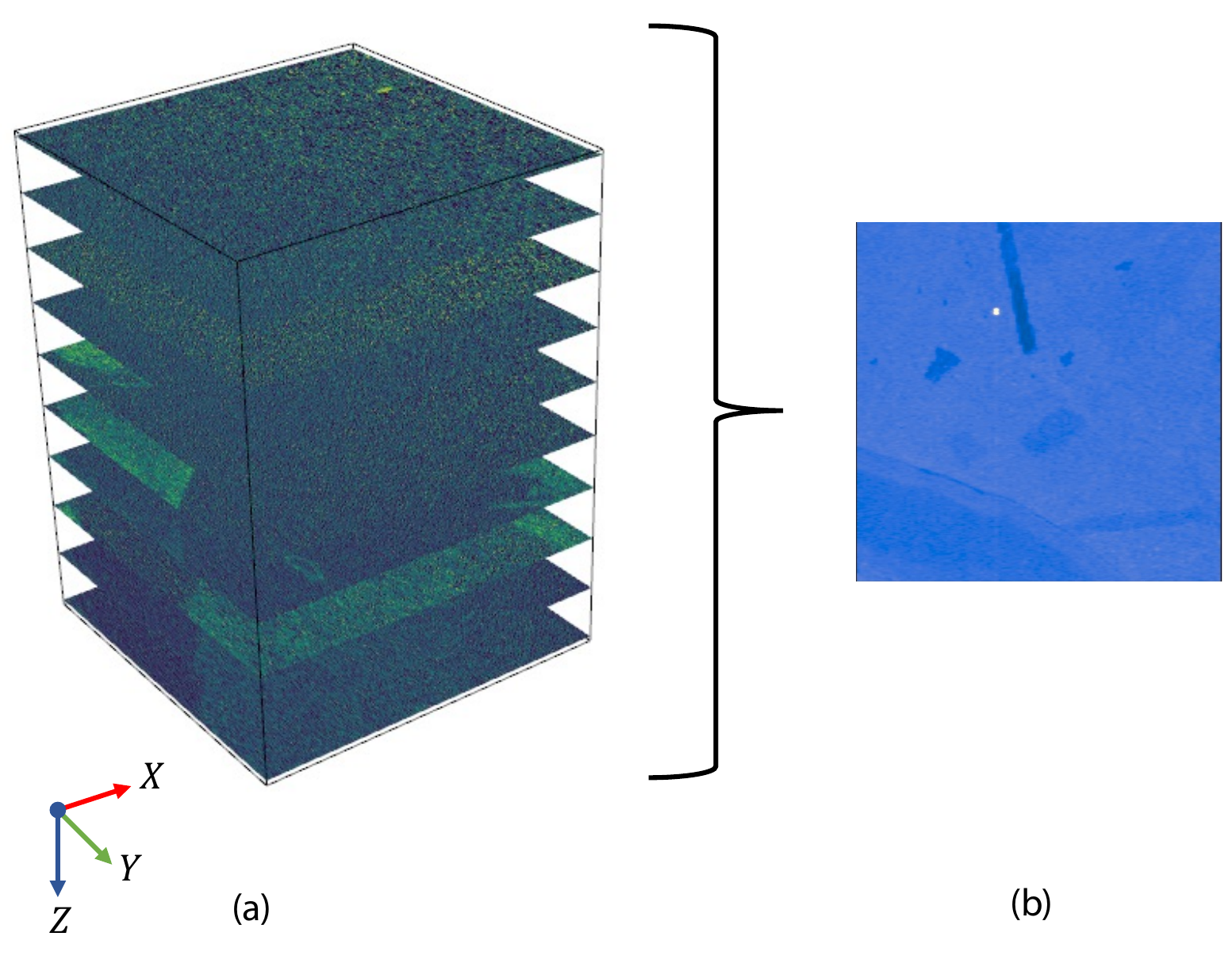}
    \caption{Applying an operation on the A-scans of the volume to create Axial Projection Image.}
    \label{fig:zslice}
\end{figure}

As proposed in \cite{Weiss2020}, axial projection maps, generated by applying an operation on the iOCT A-scans of the volume, can result in feature maps with strong visual cues for locating an instrument in the volume.
Leveraging the instrument shadowing generated by attenuation of the emitted iOCT spectrum at the needle surface, we apply an operation on the A-scans to highlight the needle footprint on the generated projection image. 
According to \cite{Weiss2020}, we name this image the \textbf{Axial Projection Image}, which can be defined as:
\begin{equation}
    \mathcal{P}_{Axial} = \bigotimes^Z \mathcal{V}[..., i]
\end{equation}
where $\mathcal{V}$ is the original volume and $\bigotimes$ is an operator. 
We choose $\bigotimes$ to be the \textit{mean operator}, as it has shown promising results in our experiments.
Visual examples of the generated projection images can be seen in Figure~\ref{fig:zslice}.

\begin{figure}[ht!]
    \centering
    \includegraphics[width=1\linewidth]{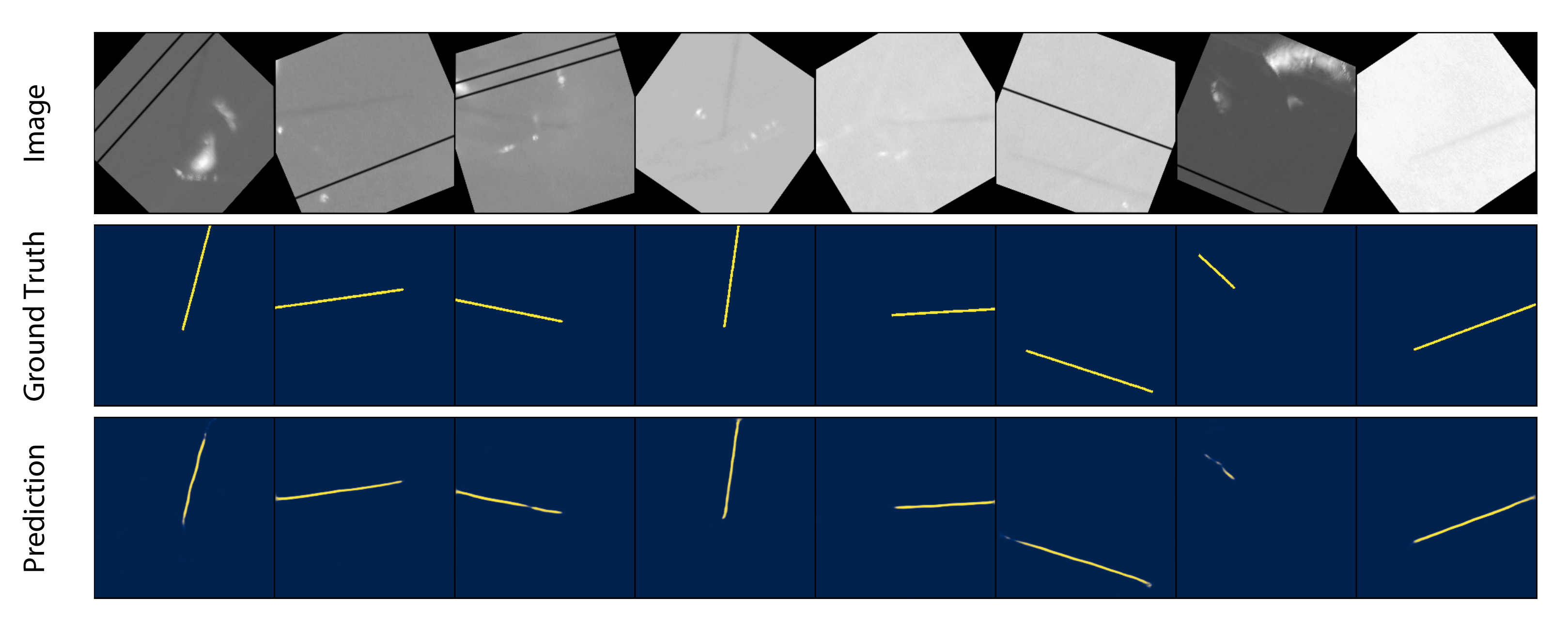}
    \caption{Performance of the Axial Image Segmentation model on the test set. Shown ground truths are including the augmentations.}
    \label{fig:enfaceresnet}
\end{figure}

By employing a CNN-based approach to segment the instrument in the 2D projection, all A-scans containing the instrument can be identified in the volume. 
Alike \cite{Weiss2020}, we train a U-Net \cite{Ronneberger2015} with ResNet18 \cite{he2015deep} backbone for the segmentation with Binary Cross Entropy Loss on our custom dataset, $\text{ds}_\text{axial}$ (Sec.~\ref{sec:materials}).
Examples of the predicted instrument segmentation and the corresponding ground-truth in our test set are shown in Fig.~\ref{fig:enfaceresnet}.

During inference, we filter the pixels with high confidence and fit a 2D line to them using Huber Regressor \cite{huber1977robust}. 
During our experiments we chose a confidence value based on the highest 1\% output probabilities.
We name the rotation of this line as $\theta_z$ (Fig.~\ref{fig:bscangeneration} (c)), as this value encodes the needle rotation w.r.t. the volume's $Z$-axis.
The needle tip position in this image also provides $t_x$ and $t_y$, defining the needle tip translation in volume $X$ and $Y$ axes, being defined as the coordinates of the extremum point on the line. 

\subsection{Intelligent Virtual B-scan}
\label{sec:MethodBScan}

\begin{figure}[ht!]
    \centering
    \includegraphics[width=1\linewidth]{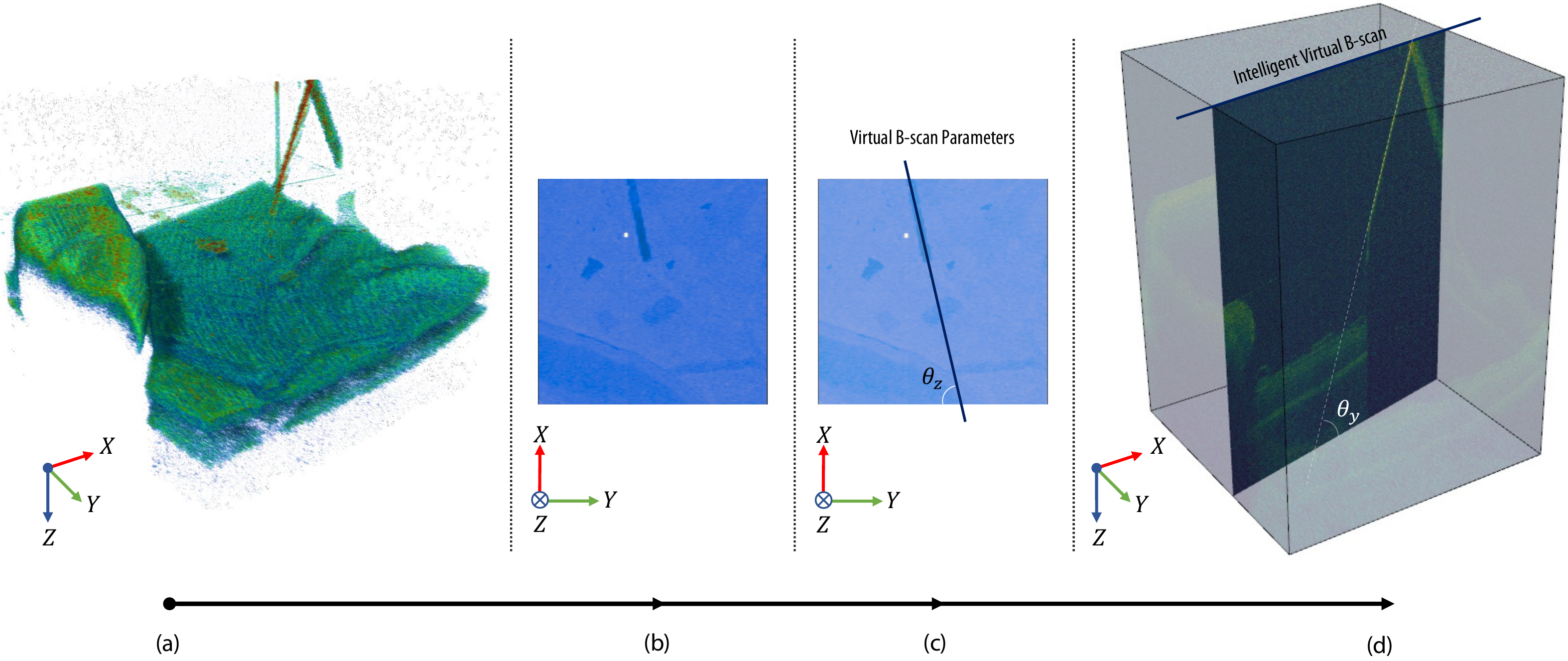}
    \caption{The method of the \textit{intelligent virtual B-scan} generation. (a) Original volume. (b) Axial Projection Image, generated with a $mean$ operation. (c) Fitted line on the segmentation result for estimating the virtual B-scan parameters. (d) \textit{Intelligent virtual B-scan}, generated from interpolation between adjacent A-Scans.}
    \label{fig:bscangeneration}
\end{figure}

The location of each B-scan in the volume can be identified by the conventional plane equation in 3D space, $(P_0, \vec{n})$, where $P_0$ is a 3D point on the plane and $\vec{n}$ is the plane normal vector.
Hence, the plane formulation of each of the B-scans acquired from the microscope can be written as:
\begin{equation}
\label{Eq:Bscans}
(P_0:(0, i \times \xi, 0), \vec{n}:\vec{j}) \\
\end{equation}
where $\xi$ is the spacing in $Y$-axis, $i$ is the index of the specific B-scan and $\vec{j}$ the unit vector in $Y$-axis.
Contrary, we define \textit{virtual B-scan}, as an arbitrary plane in the original iOCT volume. 
Instead of extracting the B-scan cross-sections provided by the iOCT scanning pattern from the volume, any arbitrary volume slicing can be defined by a plane equation.
We call a slicing \textit{intelligent}, if the plane is automatically selected based on a semantic understanding of the surgical scene and the application, in order to create an optimized representation of complex data.
The pixel intensities of this slice are composed by interpolating the voxel intensities adjacent to the virtual plane.
In our example, we use the instrument line parameters $\theta_z$, $t_x$ and $t_y$ from \ref{Sec:MethodAxial} to define a tool-aligned intelligent virtual B-scan as:
\begin{equation}
  ((t_x, t_y, \cdot), \langle\sin{\theta_z}, \cos{\theta_z}, 0\rangle)  
\end{equation}
which provides a plane aligned with the needle.
The tool-aligned slicing can therefore be generated by interpolation of the adjacent iOCT A-scans, identified by the line in Sec.~\ref{Sec:MethodAxial} (Fig.~\ref{fig:bscangeneration}). 
While in general, for the composition of the \textit{intelligent virtual B-scan}, a trilinear or more complex interpolation method could be chosen to generate a visually appealing cross-section, we employ a linear interpolation scheme to reduce the computational complexity.

\begin{figure}[h!]
    \centering
    \includegraphics[width=1\linewidth]{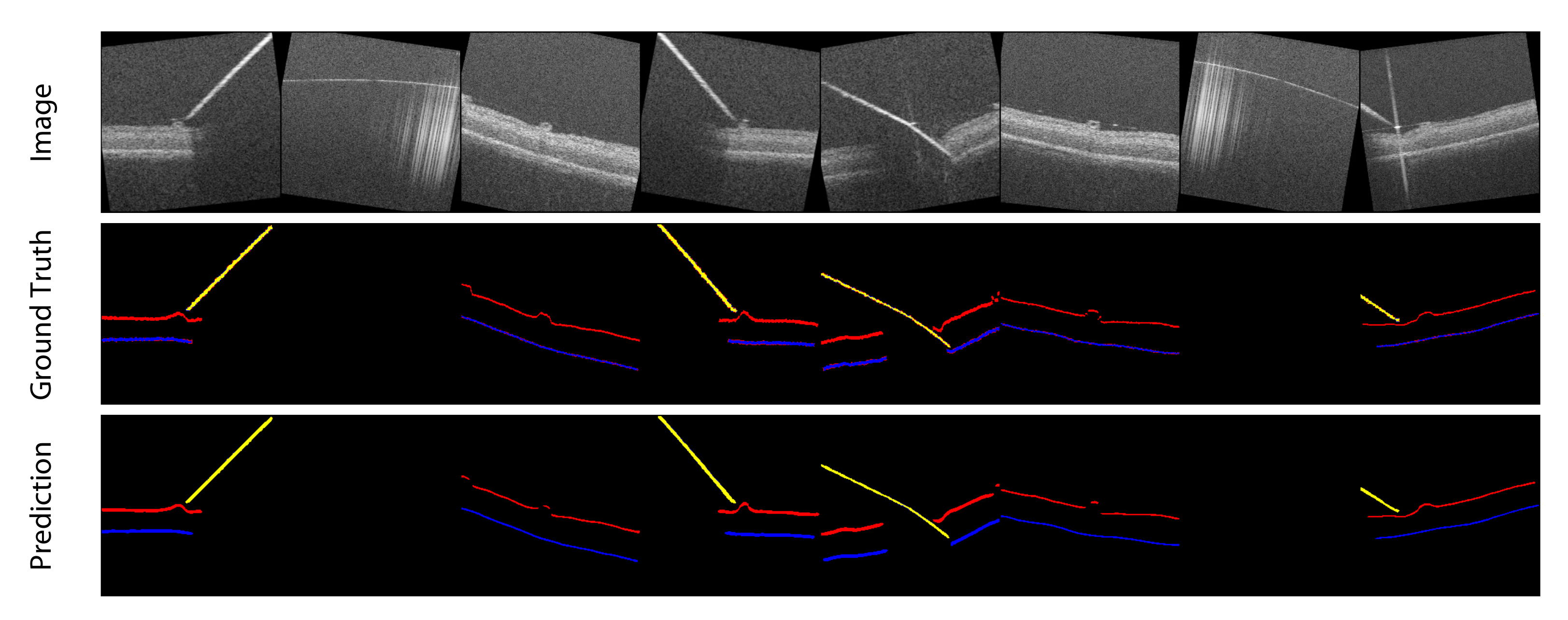}
    \caption{Example of the B-scan Segmentation model on the test data. Yellow color represents the needle, red represents ILM layer and blue, the RPE layer.}
    \label{fig:segmentationresnet}
\end{figure}

This tool-aligned \textit{intelligent virtual B-scan} allows the estimation of the axial instrument rotation $R_y$ and the instrument tip coordinate in the volume $Z$-axis, $t_z$.
To identify the needle, Inner Limiting Membrane (ILM) and RPE layers, we subsequently employ a second segmentation network.
Using our custom dataset, $\text{ds}_\text{bscan}$ (see Sec. \ref{sec:materials}), we train a U-Net style network with ResNet18 backbone, using a combination of Cross Entropy Loss and Focal Loss for the joint segmentation of retinal layers and surgical instrument, as proposed by \cite{Sommersperger2021}. 
The segmentation network was additionally pre-trained on synthetic iOCT data generated by the method described in \cite{Sommersperger2022}.
Examples of the \textit{virtual B-scans} and the obtained segmentations are illustrated in Fig.~\ref{fig:segmentationresnet}.

Similar to \ref{Sec:MethodAxial}, we fit a line to the segmented needle pixels with high confidence. We name the rotation of this line $\theta_y$ (Fig.~\ref{fig:bscangeneration} (d)), since it provides the needle $R_y$. Furthermore, $t_z$ can also be estimated from the innermost segmented point on the line.
Having $R_z$, $R_y$, $t_x$, $t_y$ and $t_z$ calculated, the 5-DoF needle pose can be formulated as:

\begin{equation}
\label{eq1}
R= R_z(\theta_z)  R_y(\theta_y)  R_x(\cdot)
\end{equation}
assuming  yaw, pitch and roll rotation around the $Z$, $Y$ and $X$ axes. Hence, we obtain the following rotation matrix:
\begin{equation}
R=\begin{pmatrix}
\cos{\theta_{z}} & -\sin{\theta_{z}} & 0 \\
\sin{\theta_{z}} & \cos{\theta_{z}} & 0 \\
0 & 0 & 1\\
\end{pmatrix}
\begin{pmatrix}
\cos{\theta_{y}} & 0 & \sin{\theta_{y}} \\
0 & 1 & 0 \\
-\sin{\theta_{y}} & 0 & \cos{\theta_{y}}
\end{pmatrix} I_3 \\
\end{equation}
In this case, the rotation around $X$-axis (roll) can be neglected due to the needle symmetry in this case.
For the tooltip position, we obtain:
\begin{equation}
t = [t_x, t_y, t_z]
\end{equation}

\subsection{Robot to Volume Registration}
\label{sec:MethodRegistration}
In iOCT-guided interventions, the scanning area can be dynamically adjusted by the surgeon.
Therefore, pre-operational calibration is not valid as the scanning region is updated, and a dynamic online registration between the robotic system and the iOCT volume is preferred.
In our case the $Z$-axis of the robot base and iOCT $Z$ directions are aligned, being parallel to the A-scan direction of the iOCT system.
To estimate the transformation between the robot and volume, it is therefore sufficient to estimate the transformation of one of the other two axes.
As the injection needle is rigidly mounted on the robot, the robot axes can be defined as:
\begin{equation}
\vec{v_z}= -\vec{k}, \quad \vec{v_y} = \langle\sin{\theta_z},\cos{\theta_z}, 0\rangle, \quad \vec{v_x}= \vec{v_y} \times \vec{v_z}
\end{equation}
Given these vectors, any translation vector, $T_{v}$, in the volume coordinate system can be transformed into a translation vector in the robot coordinate system as:
\begin{equation}
\label{eq:calibration}
  C = 
  \left[ {\begin{array}{ccc}
   \vertbar & \vertbar & \vertbar \\
   v_x & v_y & v_z \\
   \vertbar & \vertbar & \vertbar
  \end{array} } \right]\\
\end{equation}
\begin{equation}
      T_{r} = C^{-1} T_{v}
\end{equation}

Eq.~\ref{eq:calibration} transforms the translation vectors in the next step, from iOCT volume coordinates to robot coordinates.

\subsection{Trajectory Planning}
\label{sec:TrajectoryPlanning}
We define \textit{insertion line}, similar to \cite{dehghani2022colibridoc}, which is a 3D line through the target point, $V$ and parallel to the needle.
In order to execute a successful insertion, from the time of contact between the tissue and tool, the needle needs to follow this line to reach the target point and perform a successful injection.
We decouple the trajectory into two parts: \textbf{a)} $t_A$, a translation to align the needle with the \textit{insertion line} and \textbf{b)} $\vec{t_B}$, a translation along the \textit{insertion line} until the target point $V$ is reached.

\begin{figure}[H]
\label{fig:trajectory}
    \centering
    \includegraphics[width=0.8\linewidth]{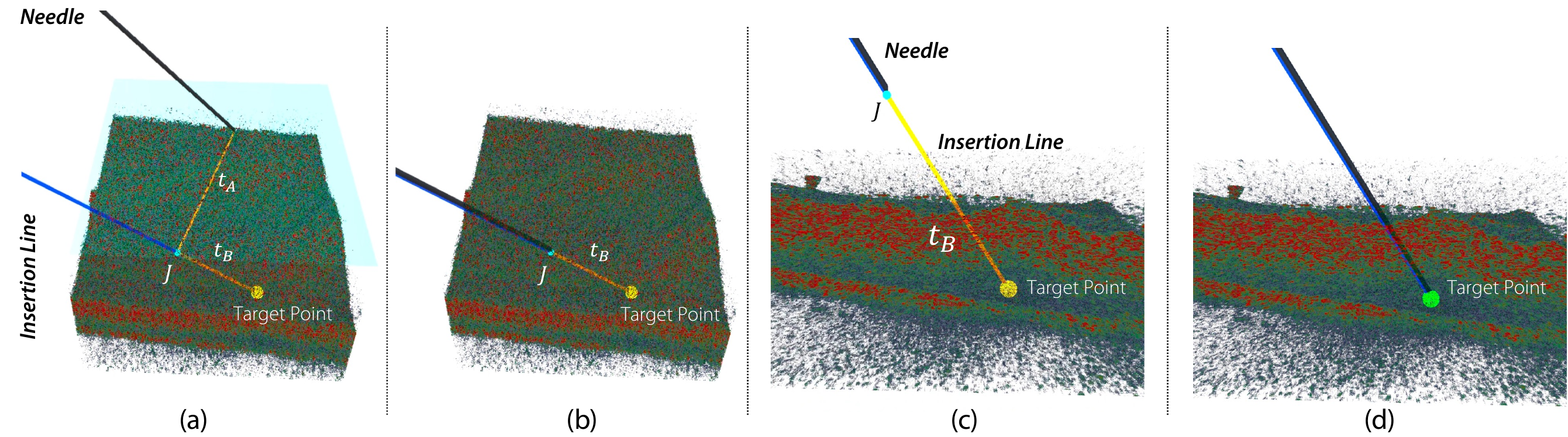}
    \caption{Steps of trajectory calculations. (a) Original volume with the target point, needle, injection line, $t_A$ and $t_B$ (b) Visualization of the volume, needle and target point after applying $t_A$ (c) Side view of previous step (d) Side view of the volume, needle and target point after applying $t_B$}
\end{figure}

There are multiple possibilities to define $\vec{t_A}$ and $\vec{t_B}$. 
However, in order to avoid tissue damage while applying $t_A$, we apply these constraints on $t_A$ and $t_B$:
\begin{equation}
\label{eq:trajectory}
\begin{cases}
    \vec{t_A} + \vec{t_B} = V - t_{needle}, \\
    \vec{t_B} \parallel r_{needle}, \\
    \angle (\vec{t_A}, \vec{k}) = \frac{\pi}{2}
\end{cases} 
\end{equation}
We calculate point $J$, as the intersection of plane $(t_{needle}, \vec{k})$ and the \textit{insertion line}. 
Thereafter, $\vec{t_A} = J - t_{needle}$ and $\vec{t_B} = V - J$, as illustrated in Fig.~\ref{fig:trajectory}.

To formulate these equations, only the \textit{intelligent virtual B-scan} described in Sec.~\ref{sec:MethodBScan} containing the needle, and a second virtual B-scan, which contains the needle after applying $t_A$, are required. 
The second \textit{virtual B-scan} can be identified as $(V, \langle\sin{\theta_z}, \cos{\theta_z}, 0\rangle))$. 
Since in our iOCT system the cross-sections include different refractive indices along the A-scans when the emitted iOCT spectrum traverses different media, the acquired points need to be compensated for the displacements in $Z$-axis \cite{turani2018refractive}.
We leverage the segmented ILM and RPE layer boundaries from the second virtual B-scan to correct the axial spacing by applying the appropriate refractive index to each of the regions, since the A-scan fixation is only applicable to the step $t_B$.

\section{Materials}
\label{sec:materials}
For our experiments we used a setup consisting of \textit{Steady Hand Eye Robot}\cite{he2012toward} and a \textit{Leica Proveo 8 With EnFocus OCT Imaging System} (Fig.~\ref{fig:setup}).
Both robot and microscope are connected to a workstation with LAN connections through a network switch, establishing the communication between all components. 
The microscope was pre-calibrated, aligning the iOCT beam to the surgeon’s view and the video signal to the iOCT scanning area.
To acquire the iOCT datasets for our experiments and network training we collected B-scans each composed of 1000 A-scans with an axial resolution of 1024 pixels each.
Each volume was collected at a squared scanning area of $2.5mm \times 2.5mm$ at a scanning depth of $3.38mm$, containing 100 linearly acquired B-scans.
This setup for the iOCT volumes provides a voxel spacing of  $(2.5 \mu m, 25 \mu m, 3 \mu m)$.
For $\text{ds}_\text{axial}$ in Sec.~\ref{Sec:MethodAxial}, we have created a dataset of Axial Projection Images from 100 iOCT volumes, as shown in Fig.~\ref{fig:enfaceresnet}. 
While acquiring the iOCT volumes from the microscope during inference time, some B-scans are not successfully transmitted due to network connection issues.
This can decrease the instrument segmentation performance, because the missing B-scans generate a similar effect in the Axial Projection Image as the needle shadowing. 
In this order, we randomly omit some of the B-scans during training.
For the  $\text{ds}_\text{bscan}$ described in Sec.~\ref{sec:MethodBScan}, we have collected a custom dataset with 300 images, $2.5 mm \times 3.8 mm$, using a \textit{NanoFil NF36BV 36 Gauge Bevealed Needle} and an \textit{ICSI MIC-SI-0 Micropipette}, above and inserted into ex-vivo pig eye retinas, in absence and presence of vitreous.
We have only collected B-scans which were obtained from the microscope directly while having the B-scans aligned with the needle.
We will make both datasets publicly available soon.

All of the CNN networks were trained using Pytorch 1.10.1 framework, using NVIDIA RTX 2080Ti.
The datasets were labeled by two biomedical engineering experts using ImFusionLabels (ImFusion GmbH, Munich, Germany). 
The acquired iOCT volumes are visualized on the workstation via a custom rendering plugins integrated in the ImFusionSuite\footnote[1]{https://www.imfusion.de} framework. (ImFusion GmbH, Munich, Germany).

To correct the pixel spacing resulted by deflection in Sec.~\ref{sec:TrajectoryPlanning}, we used $n_\text{air} = 1$, $n_\text{vitreous} = 1.38$ and $n_\text{tissue} = 1.38$, as suggested by the microscope manufacturer, during open-sky \textit{ex-vivo} porcine eye experiments.

\section{Experiments and Results}

\subsection{Inference Rate}
With inference on an NVIDIA GeForce RTX 2080Ti, for the step of \textit{intelligent virtual B-scan} composition and segmentation with needle pose estimation we measured average inference time of $0.232 \pm 0.008$ seconds (step \textbf{a} in workflow Fig.~\ref{fig:workflow}), and $0.188 \pm 0.009$ seconds for trajectory planning (step \textbf{b}).
We measured $5.45 \pm 0.80$ seconds for the robot to perform the needle insertion (step \textbf{c}), averaged over 10 insertions in randomly selected positions.
With an average image acquisition time of $7.69 \pm 0.27$ seconds, this step has shown to be the most time consuming step, as our microscopic system has not been specifically designed for integration in a robotic setup.

\subsection{System Accuracy}
\begin{figure}[ht]
    \centering
    \includegraphics[width=1\linewidth]{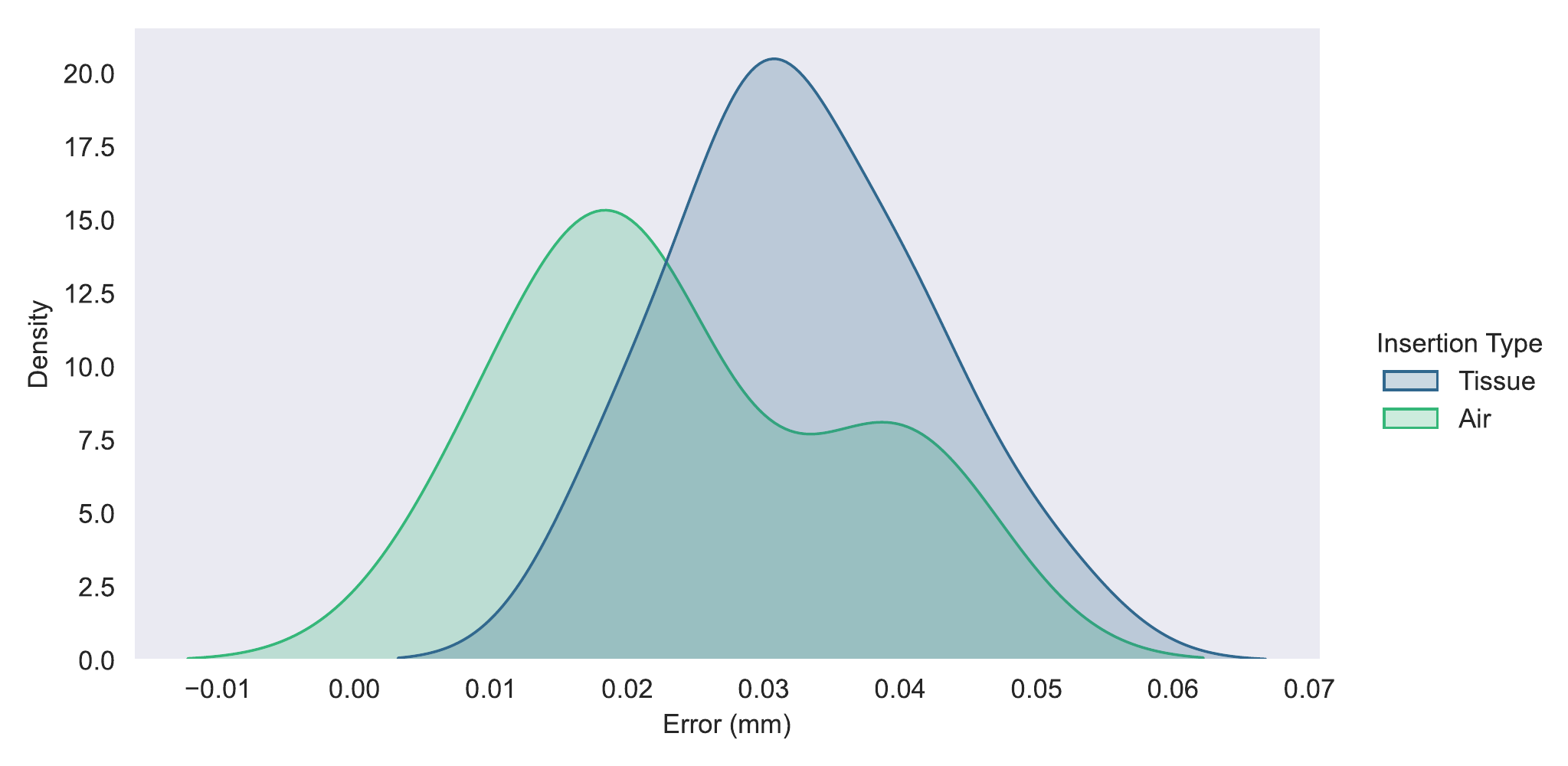}
    \caption{Final performance of the injection process. The chart shows the Euclidean distance between the needle tip and the target point, in two different environments.}
    \label{fig:evaluation}
\end{figure}

We individually evaluated the precision of the robot, the precision of the system in absence of tissue and the precision of autonomously navigating to an injection target in the retina, using the needles described in Sec.~\ref{sec:materials}.
We have not reported the accuracy of the segmentation models individually, since the performance of these networks has been evaluated and discussed for similar use cases in the referenced works.
Two biomedical engineering experts independently marked the tool tip locations in all the volumes, to generate a reliable ground truth position. 
To measure the accuracy of the robot's translational movement, we moved each robot axis about $500 \mu m$ for 10 times and measured the Euclidean distance of the needle tip in the iOCT volume before and after each movement. 
The result proves an accuracy of $5 \pm 2 \mu m$ for each of the axes of the robot. 
To measure the precision of autonomous robot navigation, we first performed 10 trials of navigating to a target location without retinal contact.
Afterwards, we performed 10 autonomous needle insertions into the retinal tissue, chose a random target point in proximity to the RPE.
For these experiments we used ex-vivo porcine eyes in an open-sky setup without removing the vitreous gel.
For all trials we measured the Euclidean distance of the needle tip's final position to the target point. 
For the 10 trials without retina insertion we measured a mean error of $24 \pm 5 \mu m$, while for navigation to a target inside the retina we measured an average error of $32 \pm 4 \mu m$.
The error distribution for both trials is illustrated in Fig. \ref{fig:evaluation}.

\section{Discussion and future work}
\label{sec:discussion}
In this work we introduced an approach for autonomous robot navigation based on volumetric iOCT tailored for subretinal injection.
We have proposed an end-to-end framework that integrates a robotic platform as well as an iOCT system, while keeping the surgeon in loop.
Our experiments on ex-vivo porcine eyes show promising results with errors close to the precision, that is required for targeting the anatomical area for subretinal injection \cite{Karampelas1256}.
The comparison to our experiments in air suggests that tissue and instrument deformations need to be investigated in future works.

In this paper, we introduce \textit{intelligent virtual B-scan} composition to reduce the complexity of the scene and focus the processing on the relevant structures of the volume.
Such representations not only allow rapid estimation of the instrument pose, but may also be displayed to the surgeon to verify the processing algorithms, which in turn could accelerate the acceptance of autonomous systems.
During our experiments we used a straight needle. However, when using a bent needle or other micro-surgical tools, where the instrument cannot be captured by a single 2D plane, multiple cross-sections could be extracted and a similar method with modifications on Sec.~\ref{Sec:MethodAxial} and Sec.~\ref{sec:MethodBScan} could be employed.
To understand the feasibility of such an end-to-end system, we restricted the robot control for our open-sky experiments and only applied translational movements.
In a realistic scenario, rotations around the remote center of motion (RCM) need to be applied to minimize the force on the sclera.
However, our method can be adapted to integrate RCM-based rotations by adjusting $t_A$ in Sec.~\ref{sec:TrajectoryPlanning}, with series of translational and rotational movements.
In a surgical scenario, patient breathing and instrument navigation can lead to significant retinal motion, therefore a higher volume acquisition rate in state-of-the-art swept-source iOCT systems can facilitate the integration of temporal registration, which in turn can be used to track the target point and update the trajectory accordingly.
Another possible extension is dynamically aligning the iOCT scanning area with the generated tool-aligned B-scans and hence, adjusting the iOCT scanning pattern to only acquire selected B-scans, which will increase the overall inference speed of our end-to-end system.
Currently, we have not yet optimized the inference times of our network, since the current bottleneck of our system is the acquisition and transmission of the iOCT volumes.
However, previous works \cite{Borkovkina2020,Sommersperger2021,Weiss2020} have shown that network inference times for iOCT segmentation can be drastically improved using kernel optimization and tensor fusion strategies.
Additionally, to further improve the real-time capability of such systems, the communication between the iOCT and robotic system may be optimized using faster communication protocols.
In future works, we will consider the segmented ILM and RPE to safely navigate the instrument, using real-time processing as proposed by \cite{Sommersperger2021}, to avoid harming RPE and cells of the neuroretina. We chose a fixed speed for the robot during trajectory execution. However, the navigation speed could be optimized by adopting a higher speed for movement without tissue contact during $t_A$ and a dynamic speed for $t_B$ to ensure effective tissue insertion.

\section{Conclusion}
\label{sec:conclusion}
In this paper we proposed an end-to-end framework for robotic subretinal injections and have designed a workflow with the surgeon in the loop, which is also adoptable to future extensions.
We proposed a fast processing method for volumetric iOCT data to automate the navigation of the robot. 
Despite current hardware limitations for real-time scan acquisition, we have shown our method is capable of real-time processing, solving the 3D instrument pose estimation by relaxing it to two separate 2D segmentation tasks.
This step is mainly enabled by the composition of \textit{intelligent virtual B-scans}.
Our method forms a base for many future works based on volumetric iOCT processing, and can provide a real-time approach for similar tasks, which involve 3D pose estimation and trajectory planning for robotic retinal surgery.

\section{Acknowledgment}
We thank Leica Microsystems GmbH (Wetzlar, Germany) for their assistance with Leica Proveo 8 With EnFocus OCT, which made it possible to conduct real-time access to the microscope data.
\clearpage

\bibliographystyle{IEEEtran}
{\scriptsize
\balance
\bibliography{References}
}
\end{document}